\documentclass[conference]{IEEEtran}
\IEEEoverridecommandlockouts
\usepackage{cite}
\usepackage{amsmath,amssymb,amsfonts}
\usepackage{algorithmic}
\usepackage{graphicx}
\usepackage{textcomp}
\usepackage{booktabs}
\usepackage{gensymb}
\usepackage{booktabs}
\usepackage{graphicx}
\usepackage[table,xcdraw]{xcolor}
\usepackage[normalem]{ulem}
\useunder{\uline}{\ul}{}
\def\BibTeX{{\rm B\kern-.05em{\sc i\kern-.025em b}\kern-.08em
    T\kern-.1667em\lower.7ex\hbox{E}\kern-.125emX}}
\begin{document}

\title{Precision Agriculture Revolution: Integrating Digital Twins and Advanced Crop Recommendation for Optimal Yield\\
}


\author{
\IEEEauthorblockN{
Sayan Banerjee\IEEEauthorrefmark{1}, 
Aniruddha Mukherjee\IEEEauthorrefmark{1},
Suket Kamboj\IEEEauthorrefmark{1} 
}
\IEEEauthorblockA{\IEEEauthorrefmark{1}Computer Science and Engineering\\
KIIT University\\
Bhubaneswar, Odisha, India\\
Email: \textbf{\texttt{\{21051087,2205533,2205248\}}}@kiit.ac.in}
}

\maketitle

\begin{abstract}
With the help of a digital twin structure, Agriculture 4.0 technologies like weather APIs (Application programming interface), GPS (Global Positioning System) modules, and NPK (Nitrogen, Phosphorus and Potassium) soil sensors and machine learning recommendation models, we seek to revolutionize agricultural production through this concept. In addition to providing precise crop growth forecasts, the combination of real-time data on soil composition, meteorological dynamics, and geographic coordinates aims to support crop recommendation models and simulate predictive scenarios for improved water and pesticide management.
\end{abstract}


\section{Objectives}
\begin{itemize}
\item Data Fusion: Consolidate and collect data on soil composition, exact location, and changing environmental conditions by integrating NPK sensors, GPS modules, and weather APIs.
\item Enhanced Models: By including multi-dimensional data inputs, crop recommendation models are improved and crop selection is more accurate and environmentally conscious.
\item Digital Twin Framework: Create a digital twin framework to simulate various crop development scenarios. This will enable careful resource management, particularly with regard to the application of pesticides and water.
\item Predictive Precision: Using advanced environmental analysis and simulated crop growth patterns, forecast the optimal crop production and harvest time.
\end{itemize}

\section{Beneficiaries}
This paper aims to benefit a wide range of stakeholders in the agriculture industry. Farmers will benefit greatly from having improved decision-making skills based on current weather patterns and soil properties, which will allow them to choose crops with the best possible yields. Better agricultural techniques would benefit agribusinesses by increasing production and decreasing resource waste, which would ultimately boost profitability. Comprehensive data would help agronomists make well-informed recommendations on sustainable crop choices and soil management. Agricultural technology developers would gain knowledge that would help them improve inventions and produce more sophisticated and effective farming implements. 

Additionally, programs that prioritize environmental sustainability would value the effective use of resources that these results encourage, which is consistent with conservation goals. Increased agricultural output would lead to stabilized supply chains in the agriculturally dependent economic sectors, which would help the food industry and related businesses. All things considered, this study provides a thorough framework that incorporates real-time data from multiple sources, enabling stakeholders to make precise decisions and advancing agricultural sustainability.

\section{Value of Results}
\begin{itemize}
\item Optimized Decision-making: Enhanced crop selection based on current weather patterns and soil properties for optimal decision-making.
\item Resource Efficiency: Reduced costs and ecological sustainability through simplified resource management such as pesticides, insecticides and water.
\item Accurate Forecasting: Accurate estimates of crop growth phases for better planning and yield estimation.
\end{itemize}

\section{Background}
The introduction of new technologies has caused a considerable change in traditional farming techniques. An era of data-driven decision-making in agriculture has begun with the adoption of modern technologies into farming techniques, or ''Agriculture 4.0". With the advent of technologies like GPS modules, weather APIs, and NPK soil sensors, farmers now have invaluable access to information about weather patterns, soil health, and environmental conditions.
Farmers can precisely map their fields and understand spatial variations in crop production and soil properties by using GPS modules, which enable precise geolocation. Weather APIs give farmers access to anticipated and real-time meteorological data, including vital details about temperature, humidity, and precipitation, so they may make well-informed decisions that are in line with the weather.

\begin{table*}[ht]
\centering
\caption{Summary of Research Papers on NPK Sensors, Weather APIs, Digital Twins, and Agriculture 4.0.}
\label{tab:research-papers}
\renewcommand{\arraystretch}{1.2} 
\begin{tabular}{p{0.35\textwidth}p{0.40\textwidth}p{0.15\textwidth}}
\toprule
\textbf{Paper Title} & \textbf{Key Focus} & \textbf{Publication Date} \\
\midrule
Sustainable Crop Recommendation System Using Soil NPK Sensor~\cite{rel-1} & Developing a system for crop recommendations based on soil NPK sensor data & October 2023 \\
\hline
AI-Based Real-Time Weather Condition Prediction with Optimized Agricultural Resources~\cite{rel-2} & Utilizing AI for real-time weather prediction to optimize agricultural resource management & June 2023 \\ 
\hline
Enhancing Smart Agriculture by Implementing Digital Twins: A Comprehensive Review~\cite{rel-3} & Reviewing the implementation of digital twins in smart agriculture & August 2023 \\ 
\hline
Enhancing Smart Farming Through the Applications of Agriculture 4.0~\cite{rel-4} & Describing the fourth agricultural revolution using digital technologies for smarter and more sustainable farming & August 2023 \\ 
\hline
An Overview of Agriculture 4.0 Development: Systematic Review~\cite{rel-5} & Discussing upgrades to traditional production methods through emerging technologies in the agricultural value chain & August 2021 \\ 
\hline
Agriculture 4.0: Making it Work for People, Production, and the Planet~\cite{rel-6} & Presenting solutions like AI, robotics, big data, and IoT to food production challenges & July 2020 \\ 
\bottomrule
\end{tabular}
\end{table*}

\begin{figure}[ht]
	\centering
	\includegraphics[width=0.9\linewidth]{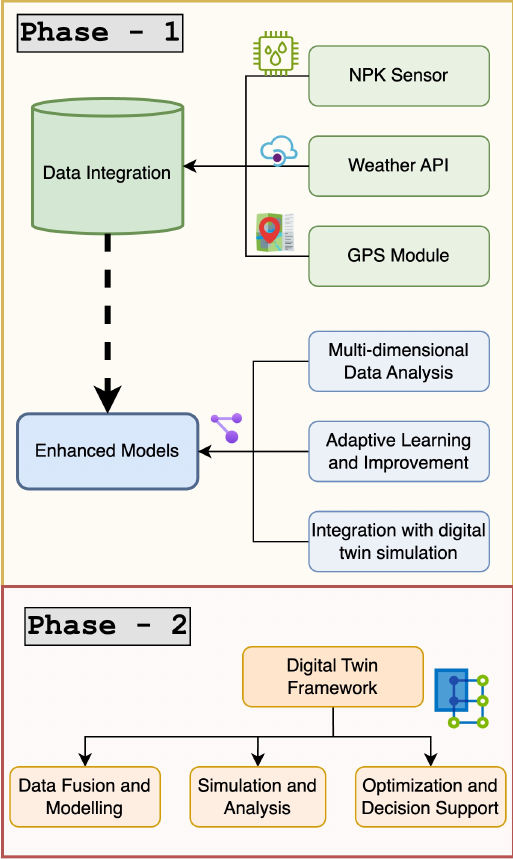}
	\caption{Block Diagram of the proposed Framework}
	\label{fig:1}
\end{figure}

NPK soil sensors are essential for determining soil fertility because they provide real-time measurements of pH, phosphorus, nitrogen, and potassium. By helping to understand nutrient excesses or deficiencies, this data enables more focused and effective fertilization techniques. Under the auspices of Agriculture 4.0, these technologies have been integrated collectively to provide farmers with useful insights that have improved agricultural productivity and optimized resource management.
Even though each of these technologies has a lot to offer on its own, there is still much to learn about how to integrate them into a coherent whole. The difference is in combining these technologies into one cohesive system that can do both predictive modeling and in-depth data analysis.

Presenting the idea of ''digital twins"—a cutting-edge technology that digitally replicates real-world assets or systems. A digital twin in the context of agriculture is a virtual version of the farm environment that incorporates real-time data inputs from several sources. With the use of this simulation framework, scenario modeling, predictive analysis, and accurate decision-making could revolutionize the agricultural industry.

The critical point is when Agriculture 4.0 technologies are combined with the capabilities of a digital twin structure, utilizing the insights obtained from weather APIs, GPS modules, and NPK soil sensors. With this combination, farmers will have a comprehensive understanding of their land, be able to accurately forecast crop growth, and be able to implement sophisticated management techniques for the most efficient use of available resources. This is an unparalleled chance to transform agricultural operations.

A comprehensive strategy that fully utilizes the potential of these technologies within a digital twin framework is important, as evidenced by the disparity between the availability of varied data streams and their integration into a predictive, actionable framework for farmers. A promising new frontier in agriculture is opened up by this integration, one that promises increased resistance to changing environmental conditions as well as increased productivity and sustainability.


\section{Statement of Problem}
Inaccurate decision-making in crop selection and resource usage is hampered by the fragmented application of Agriculture 4.0 technology, which leads to subpar yields and inefficient resource allocation.
The issue at stake concerns the fragmented application of technologies—such as GPS modules, weather APIs, and NPK soil sensors—in isolation instead of as a part of a cohesive system. The potential for thorough data analysis and predictive modeling—which are essential for accurate decision-making in agriculture—is limited by this disjointed approach. Farmers are unable to fully utilize real-time data for improved resource management and precise predictive insights in crop production due to the absence of a single framework that integrates various technologies into a coherent system.

\section{Extent of research}
Existing research focuses mostly on specific components in agriculture, such as NPK sensors, GPS modules, and weather APIs. However, the integration of these elements into a cohesive predictive framework remains relatively unexplored.

\section{Present Methods of Tackling the Problem}
Instead of having a comprehensive and predictive model, current approaches rely on using independent data sources for basic resource management and crop selection.

\section{Proposed Solution}

\subsection{Methodology}\label{AA}
\begin{itemize}

\item Data Integration:
\begin{itemize}
\item \textit{NPK Sensors:} These sensors measure the pH, phosphorus, potassium, and nitrogen levels in the soil, giving important information on the fertility and health of the soil. \\
\textbf{Precision: $\pm2\%$ FS} where FS $\hookrightarrow$ Full scale 
\item \textit{GPS Modules}: These modules pinpoint the precise geographic coordinates of the soil, allowing for accurate location-based analysis.
\item \textit{Weather APIs}: These APIs incorporate real-time weather information on temperature, humidity, and rainfall—all of which have a significant impact on crop development.
\end{itemize}

\item Digital Twin Framework:
\begin{itemize}
\item Data Fusion and Modeling: A single digital platform is created by combining the acquired data from NPK sensors, GPS units, and weather APIs. Using this platform, a digital "twin" or replica of the actual agricultural setting is produced.
\item Simulation and Analysis: A variety of crop growth scenarios are simulated using this digital twin. The system forecasts crop responses to various soil compositions, environmental factors, and geographic locations by utilizing intricate algorithms and models.
\end{itemize}

\item Optimization and Decision Support:
\begin{itemize}
\item Resource Management: By simulating crop development under various conditions, the digital twin enables predictive study of resource requirements, including water, insecticides and pesticides. It assists in figuring out when and how much irrigation and pesticide treatment are best done.

\begin{table*}[ht]
\centering
\caption{Sample Predictions}
\label{tab:my-predict}
\resizebox{0.73\textwidth}{!}{%
\begin{tabular}{|ccccccc|}
\hline
\multicolumn{1}{|c|}{N} &
  \multicolumn{1}{c|}{P} &
  \multicolumn{1}{c|}{K} &
  \multicolumn{1}{c|}{\begin{tabular}[c]{@{}c@{}}Temperature\\ (in Deg Celcius)\end{tabular}} &
  \multicolumn{1}{c|}{\begin{tabular}[c]{@{}c@{}}Humidity\\ (Relative humidity in \%)\end{tabular}} &
  \multicolumn{1}{c|}{\begin{tabular}[c]{@{}c@{}}pH of \\ the soil\end{tabular}} &
  \begin{tabular}[c]{@{}c@{}}Rainfall \\ (in mm)\end{tabular} \\ \hline
\multicolumn{1}{|c|}{90} & \multicolumn{1}{c|}{40} & \multicolumn{1}{c|}{40} & \multicolumn{1}{c|}{25} & \multicolumn{1}{c|}{82} & \multicolumn{1}{c|}{6.5} & 200 \\ \hline
\multicolumn{7}{|c|}{{\color[HTML]{000000} {\ul \texttt{Crop Recommendations}}}}                                                                                  \\ \hline
\multicolumn{7}{|c|}{{\color[HTML]{000000} Crop: rice, Probability: \textbf{0.61}}}                                                                                        \\ \hline
\multicolumn{7}{|c|}{{\color[HTML]{000000} Crop: jute, Probability: \textbf{0.39}}}                                                                                        \\ \hline
\end{tabular}%
}
\end{table*}

\item Crop Yield Prediction: The system predicts the growth timeline of crops, calculating when they will achieve maturity and be ready for harvest, by taking into account historical and current data.
\end{itemize}
\end{itemize}
\begin{figure*}[htbp]
\centerline{\includegraphics[width=0.89\linewidth]{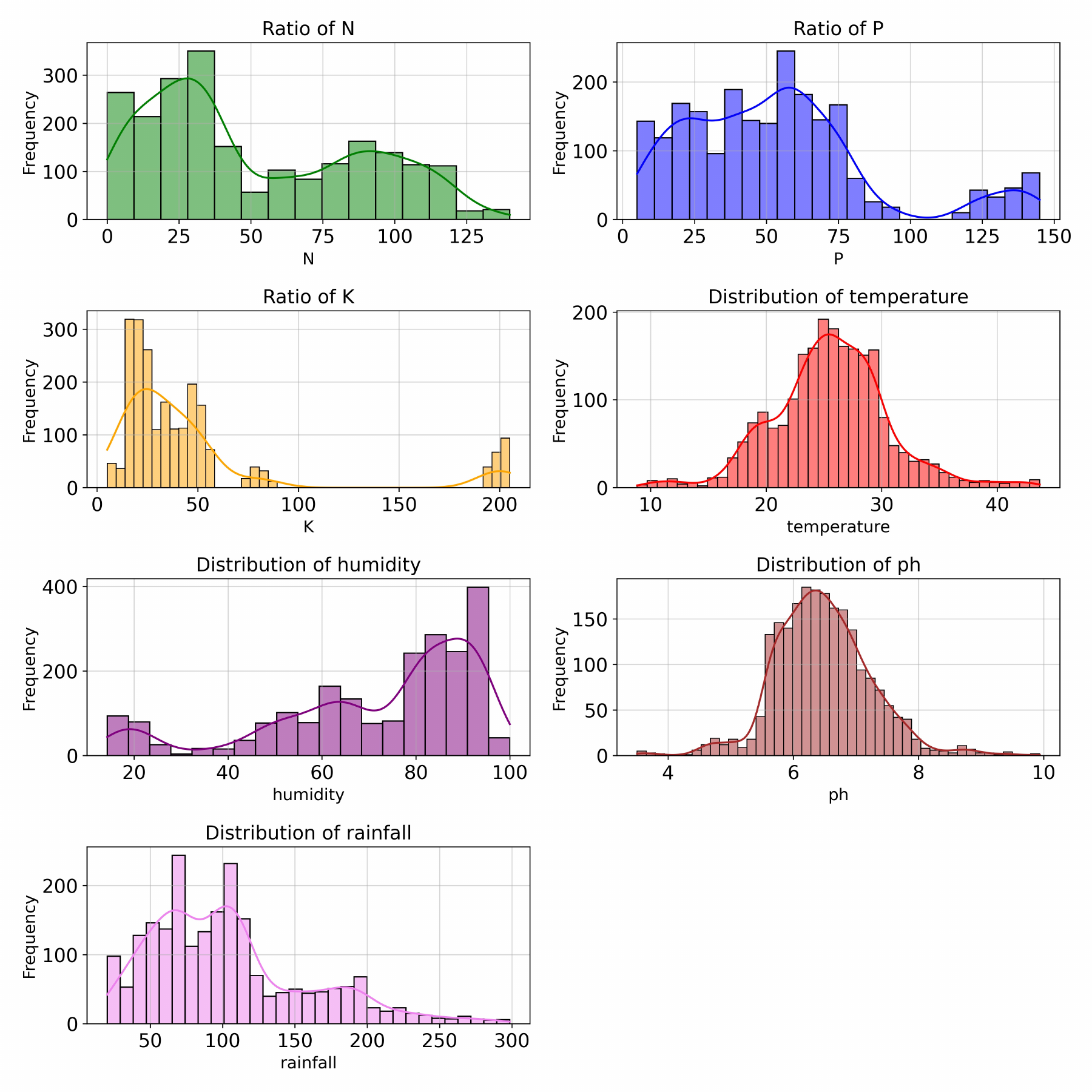}}
\caption{Correlation between different features }
\label{dataset-distribution}
\end{figure*}
\subsection{Advantages}
\begin{itemize}
\item Holistic Approach: It combines several data sources into a single, cohesive framework, enabling a thorough comprehension of the agricultural ecosystem.
\item Predictive Capabilities: By simulating various scenarios via predictive modeling, the system helps farmers make educated decisions in advance.
\item Resource optimization: It reduces water and pesticide waste by precisely forecasting resource requirements, supporting environmentally friendly agricultural methods.
\item Enhanced productivity and Efficiency: The system seeks to improve total crop productivity and operational efficiency by carefully selecting crops and utilizing resources in an optimal manner.
\end{itemize}

\section{Alternate Solution/Approaches}
Alternative approaches may encompass the utilization of machine learning algorithms for predictive modeling or the development of independent resource management optimization systems.

\section{Novelty of Approach}
\begin{itemize}
    \item Enhanced \textbf{Decision Support}: One essential component is the addition of an advanced crop suggestion tool. It processes the gathered data—soil composition, environmental factors, and geographic information—using sophisticated algorithms and data analysis to recommend the best crop kinds for particular regions.
    \item Multi-dimensional Data Analysis: This module makes use of a wide range of data inputs, such as geographic coordinates, real-time weather data, and soil nutrients (pH, nitrate, phosphorus, and potassium). Through the analysis of this multi-dimensional data, it offers precise and nuanced crop suggestions that are suited to the unique needs of the specified soil and environmental circumstances.
    \item Adaptive Learning and Improvement: The crop recommendation system continually learns and improves based on the daily real-time input from the sensors. With the help of this adaptive learning process, recommendations can be improved over time, producing choices for the best crop selection that are progressively more accurate.
    \item Integration with Digital Twin Simulation: There is a smooth integration between the crop recommendation module and the digital twin architecture. Enhancing the system's predictive skills, it makes use of its suggestions to simulate and forecast how different recommended crops will perform in various environmental circumstances.
    \item Potential for \textbf{Yield Optimization}:The module helps to maximize agricultural production potential by recommending crops that are suited for the soil and surrounding circumstances. It promotes sustainable agriculture practices, lowers the chance of crop failure, and facilitates the effective use of resources.
\end{itemize}

\section{Figures and Tables}
\paragraph{Block Diagram} It illustrates an integrated architecture for digital twin-enabled precision agriculture. Essentially, the figure presents a central node that represents the Digital Twin Framework, which integrates information from many sources such as meteorological APIs, GPS modules, and NPK soil sensors. These inputs enter the framework and start the cycle of modeling and data fusion. Based on soil and environmental parameters, crop growth predictions are produced by the simulation and analysis section of the framework. The framework then supports resource management decision-making by directing the best use of pesticides and water. The diagram shows how these interdependent elements work together harmoniously to produce educated decision-making for increased crop output and sustainable farming methods.
\\
\paragraph{Distribution graph} It shows the distribution of various features in the data-set. The x-axis on each graph represents the range of values for each variable, while the y-axis represents the number of times each value was observed. The graphs show that the most common temperatures in the city are between 20 and 40 degrees Celsius, the most common humidity levels are between 60 and 80 percent, and the most common rainfall amounts are between 50 and 100 millimeters.
\\
\paragraph{Correlation Map} The linkages between various variables are visually represented in a correlation map for a given data-set. It shows how much and in which direction these elements correlate, whether they tend to rise or fall together, or whether there is no discernible relationship between them.


\begin{figure}[ht]
	\centering
	\includegraphics[width=0.9\linewidth]{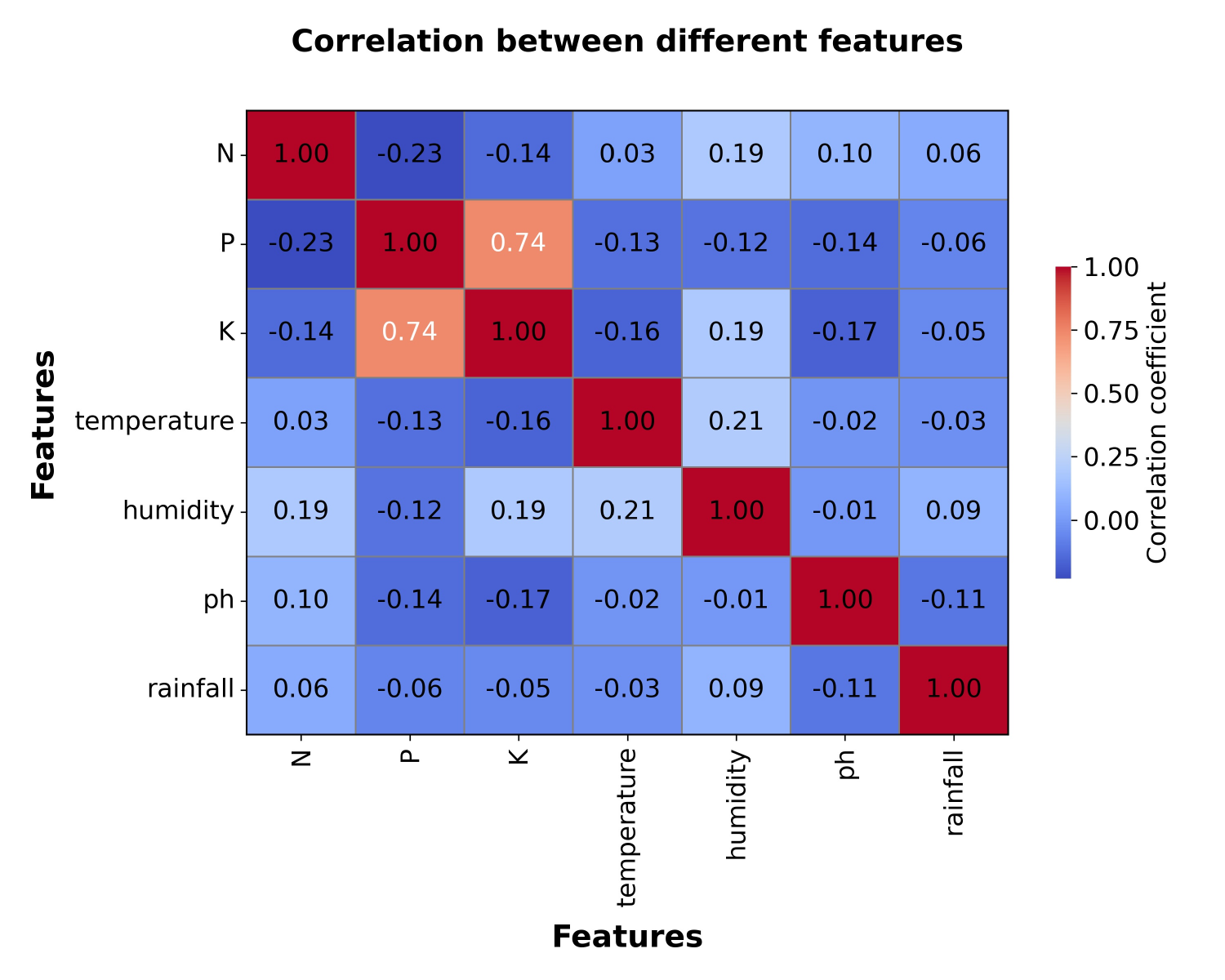}
	\caption{Correlation between different features}
	\label{heatmap_highres}
\end{figure}

Figure Labels: "Nitrogen (N), Phosphorous (P), Potassium (K), Potential of Hydrogen (pH)"
\\
\paragraph{ROC Curve} This figure depicts the receiver operating characteristic (ROC) curves for 22 classes in a multi-class classification problem.
Each curve represents the performance of the classifier for a specific class, with True Positive Rate (TPR) plotted on the y-axis and False Positive Rate (FPR) on the x-axis.
The area under the curve (AUC) is also displayed for each class, where a higher AUC indicates better classifier performance. Overall, the classifier performs well, with most ROC curves close to the top left corner of the graph, indicating excellent class distinguishability.
\begin{figure}[ht]
\centerline{\includegraphics[width=0.89\linewidth]{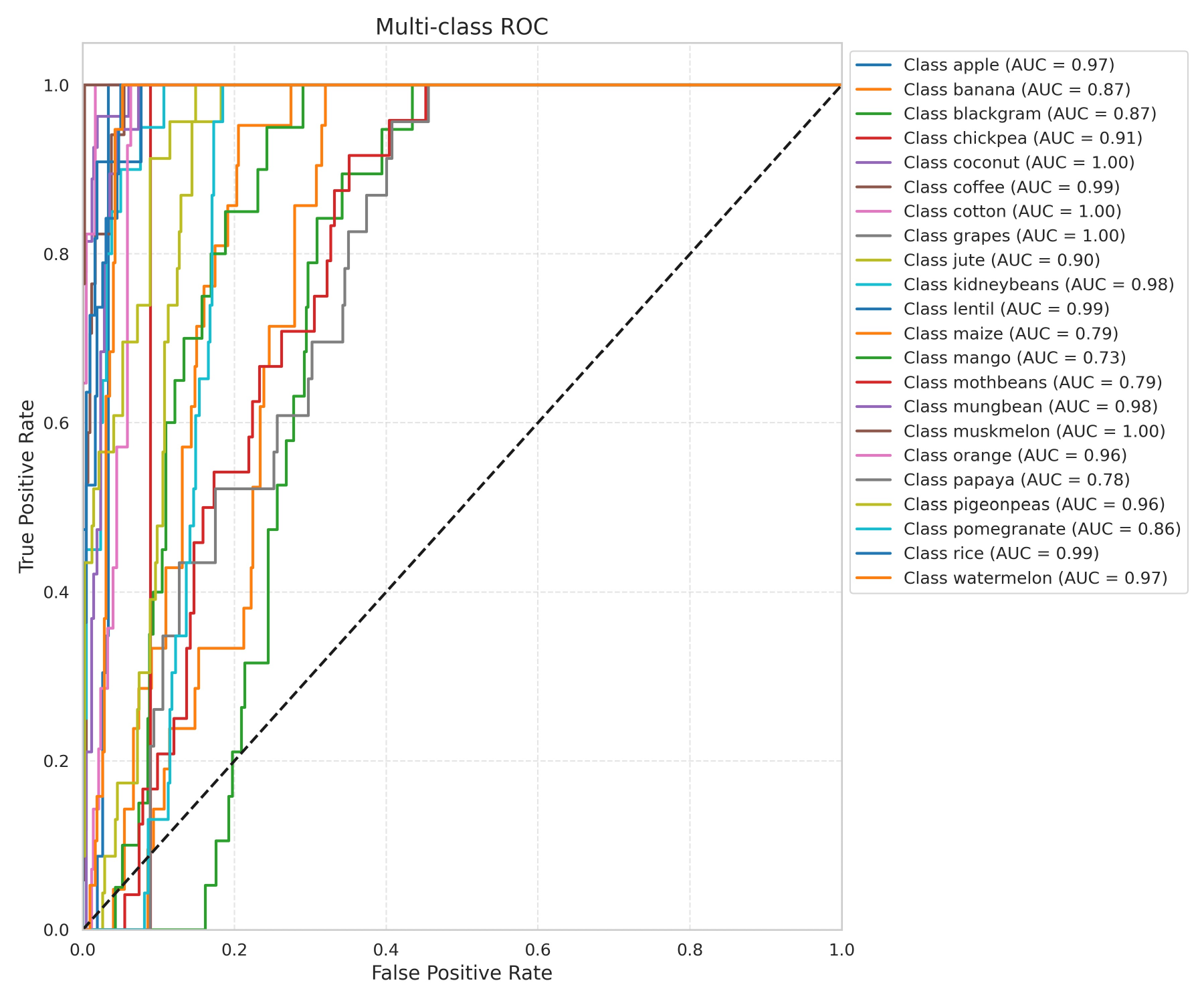}}
\caption{Multi-class Receiver Operating Characteristic graph}
\label{dataset-dist}
\end{figure}
\\
\paragraph{Metrics Table} Performance metrics for a regression model that is used for categorization are shown in the table. It comprises several parameters, including precision, recall, and F1-score for distinct crop classes (apple, banana, blackgram, etc.), along with accuracy (0.945). By taking into account the support (number of examples) for each class, these metrics assess how well the model can predict each class. Overall performance across several crop categories shows good precision, recall, and F1-scores together with excellent accuracy.

\begin{table}[ht]
\caption{Evaluation Metrics}
\centering
\begin{tabular}{lcccc} 
\toprule
\textbf{Crop} & \textbf{\textit{Precision}} & \textbf{\textit{Recall}} & \textbf{\textit{F1-Score}} & \textbf{\textit{Support}} \\
\midrule
Apple & 1.00 & 1.00 & 1.00 & 23 \\
Banana & 0.95 & 1.00 & 0.98 & 21 \\
Blackgram & 0.83 & 0.75 & 0.79 & 20 \\
Chickpea & 1.00 & 1.00 & 1.00 & 26 \\
Coconut & 1.00 & 1.00 & 1.00 & 27 \\
Coffee & 0.94 & 1.00 & 0.97 & 17 \\
Cotton & 0.80 & 0.94 & 0.86 & 17 \\
Grapes & 1.00 & 1.00 & 1.00 & 14 \\
Jute & 0.91 & 0.87 & 0.89 & 23 \\
Kidney Beans & 1.00 & 0.95 & 0.97 & 20 \\
Lentil & 0.83 & 0.91 & 0.87 & 11 \\
Maize & 0.94 & 0.76 & 0.84 & 21 \\
Mango & 0.95 & 1.00 & 0.97 & 19 \\
Moth Beans & 0.85 & 0.92 & 0.88 & 24 \\
Mung Bean & 0.95 & 1.00 & 0.97 & 19 \\
Muskmelon & 1.00 & 1.00 & 1.00 & 17 \\
Orange & 1.00 & 1.00 & 1.00 & 14 \\
Papaya & 0.95 & 0.91 & 0.93 & 23 \\
Pigeon Peas & 0.95 & 0.91 & 0.93 & 23 \\
Pomegranate & 1.00 & 1.00 & 1.00 & 23 \\
Rice & 0.89 & 0.89 & 0.89 & 19 \\
Watermelon & 1.00 & 1.00 & 1.00 & 19 \\
\\
\midrule
\textbf{Accuracy} & & & \textbf{0.95} & 440 \\
\textbf{Macro Avg} & \textbf{0.94} & \textbf{0.95} & \textbf{0.94} & 440 \\
\textbf{Weighted Avg} & \textbf{0.95} & \textbf{0.95} & \textbf{0.94} & 440 \\
\bottomrule
\end{tabular}
\label{tab1}
\end{table}
\paragraph{Summary Table} It provides a thorough synopsis of substantial research initiatives in agricultural innovation. The compilation showcases the varied range of research and emphasizes how important NPK sensor technologies, weather APIs, and digital twins are to the change of conventional farming practices. By means of comprehensive studies, the research concentrates on important areas such precision agriculture's usage of NPK monitoring, utilizing Digital Twins to overcome obstacles and improve sustainability, and putting virtual counterparts into practice to mimic actual agricultural practices. The main conclusions provide insight into the possible uses of these technologies, which include supply chain and production system optimization as well as real-time soil nutrient monitoring and customized fertilization suggestions.
\\
\paragraph{Sample Predictions} In this specific example, the table shows that with an NPK fertilizer application of 90-40-40, a temperature of 25 degrees Celsius, and a humidity of 82\%, rice is the most suitable crop to grow, with a probability of 61\%. Jute is also a potential option, but with a lower probability of 39\%.


\bibliographystyle{IEEEtran}
\bibliography{ref}

\begin{thebibliography}{10}
\providecommand{\url}[1]{#1}
\csname url@samestyle\endcsname
\providecommand{\newblock}{\relax}
\providecommand{\bibinfo}[2]{#2}
\providecommand{\BIBentrySTDinterwordspacing}{\spaceskip=0pt\relax}
\providecommand{\BIBentryALTinterwordstretchfactor}{4}
\providecommand{\BIBentryALTinterwordspacing}{\spaceskip=\fontdimen2\font plus
\BIBentryALTinterwordstretchfactor\fontdimen3\font minus \fontdimen4\font\relax}
\providecommand{\BIBforeignlanguage}[2]{{%
\expandafter\ifx\csname l@#1\endcsname\relax
\typeout{** WARNING: IEEEtran.bst: No hyphenation pattern has been}%
\typeout{** loaded for the language `#1'. Using the pattern for}%
\typeout{** the default language instead.}%
\else
\language=\csname l@#1\endcsname
\fi
#2}}
\providecommand{\BIBdecl}{\relax}
\BIBdecl

\bibitem{rel-1}
L.~Gottemukkala, S.~Jajala, A.~Thalari, S.~Vootkuri, V.~Kumar, and G.~Naidu, ``Sustainable crop recommendation system using soil npk sensor,'' \emph{E3S Web of Conferences}, vol. 430, 10 2023.

\bibitem{rel-2}
J.~Nyakuri, I.~Pierre, J.~Viviane, I.~Lambert, B.~Shadrack, N.~Erneste, N.~Schadrack, K.~Alexis, H.~Francois, and Theogene, ``Ai based real-time weather condition prediction with optimized agricultural resources,'' \emph{European Journal of Technology}, vol.~7, 06 2023.

\bibitem{rel-3}
\BIBentryALTinterwordspacing
N.~Peladarinos, D.~Piromalis, V.~Cheimaras, E.~Tserepas, R.~A. Munteanu, and P.~Papageorgas, ``Enhancing smart agriculture by implementing digital twins: A comprehensive review,'' \emph{Sensors}, vol.~23, no.~16, 2023. [Online]. Available: \url{https://www.mdpi.com/1424-8220/23/16/7128}
\BIBentrySTDinterwordspacing

\bibitem{rel-4}
\BIBentryALTinterwordspacing
R.~S. Abid~Haleem, Ravi Pratap~Singh, ``Enhancing smart farming through the applications of agriculture 4.0 technologies,'' \emph{International Journal of Intelligent Networks}, vol.~3, pp. 150--164, 2022. [Online]. Available: \url{https://www.sciencedirect.com/science/article/pii/S2666603022000173}
\BIBentrySTDinterwordspacing

\bibitem{rel-5}
\BIBentryALTinterwordspacing
F.~{da Silveira}, F.~H. Lermen, and F.~G. Amaral, ``An overview of agriculture 4.0 development: Systematic review of descriptions, technologies, barriers, advantages, and disadvantages,'' \emph{Computers and Electronics in Agriculture}, vol. 189, p. 106405, 2021. [Online]. Available: \url{https://www.sciencedirect.com/science/article/pii/S0168169921004221}
\BIBentrySTDinterwordspacing

\bibitem{rel-6}
\BIBentryALTinterwordspacing
D.~C. Rose, R.~Wheeler, M.~Winter, M.~Lobley, and C.-A. Chivers, ``Agriculture 4.0: Making it work for people, production, and the planet,'' \emph{Land Use Policy}, vol. 100, p. 104933, 2021. [Online]. Available: \url{https://www.sciencedirect.com/science/article/pii/S0264837719319489}
\BIBentrySTDinterwordspacing

\bibitem{Fuller2020May}
A.~Fuller, Z.~Fan, C.~Day, and C.~Barlow, ``{Digital Twin: Enabling Technologies, Challenges and Open Research},'' \emph{IEEE Access}, vol.~PP, no.~99, p.~1, May 2020.

\bibitem{Sharma2022Nov}
A.~Sharma, E.~Kosasih, J.~Zhang, A.~Brintrup, and A.~Calinescu, ``{Digital Twins: State of the art theory and practice, challenges, and open research questions},'' \emph{Journal of Industrial Information Integration}, vol.~30, p. 100383, Nov. 2022.

\bibitem{Wang2024Feb}
S.~Wang, J.~Zhang, P.~Wang, J.~Law, R.~Calinescu, and L.~Mihaylova, ``{A deep learning-enhanced Digital Twin framework for improving safety and reliability in human{\textendash}robot collaborative manufacturing},'' \emph{Rob. Comput. Integr. Manuf.}, vol.~85, p. 102608, Feb. 2024.

\bibitem{Javaid2022Jan}
M.~Javaid, A.~Haleem, R.~P. Singh, and R.~Suman, ``{Enhancing smart farming through the applications of Agriculture 4.0 technologies},'' \emph{International Journal of Intelligent Networks}, vol.~3, pp. 150--164, Jan. 2022.

\bibitem{Zhai2020Mar}
Z.~Zhai, J.~F. Mart{\ifmmode\acute{\imath}\else\'{\i}\fi}nez, V.~Beltran, and N.~L. Mart{\ifmmode\acute{\imath}\else\'{\i}\fi}nez, ``{Decision support systems for agriculture 4.0: Survey and challenges},'' \emph{Comput. Electron. Agric.}, vol. 170, p. 105256, Mar. 2020.

\bibitem{Rose2018Dec}
D.~C. Rose and J.~Chilvers, ``{Agriculture 4.0: Broadening Responsible Innovation in an Era of Smart Farming},'' \emph{Front. Sustainable Food Syst.}, vol.~2, p. 387545, Dec. 2018.

\bibitem{inproceedings}
M.~Masrie, M.~Rosman, R.~Sam, and Z.~Janin, ``Detection of nitrogen, phosphorus, and potassium (npk) nutrients of soil using optical transducer,'' in \emph{2017 IEEE 4th International Conference on Smart Instrumentation, Measurement and Application (ICSIMA)}, 11 2017, pp. 1--4.

\bibitem{10112795}
B.~Cheruvu, S.~B. Latha, M.~Nikhil, H.~Mahajan, and K.~Prashanth, ``Smart farming system using npk sensor,'' in \emph{2023 9th International Conference on Advanced Computing and Communication Systems (ICACCS)}, vol.~1, 2023, pp. 957--963.

\bibitem{9732000}
J.~L.~C. Ison, J.~A. B.~S. Pedro, J.~Z. Ramizares, G.~V. Magwili, and C.~C. Hortinela, ``Precision agriculture detecting npk level using a wireless sensor network with mobile sensor nodes,'' in \emph{2021 IEEE 13th International Conference on Humanoid, Nanotechnology, Information Technology, Communication and Control, Environment, and Management (HNICEM)}, 2021, pp. 1--6.

\bibitem{ALVES2023135920}
\BIBentryALTinterwordspacing
R.~G. Alves, R.~F. Maia, and F.~Lima, ``Development of a digital twin for smart farming: Irrigation management system for water saving,'' \emph{Journal of Cleaner Production}, vol. 388, p. 135920, 2023. [Online]. Available: \url{https://www.sciencedirect.com/science/article/pii/S0959652623000781}
\BIBentrySTDinterwordspacing

\bibitem{BibEntry2024Jan}
\BIBentryALTinterwordspacing
``{Detection of NPK and pH components of soil {$\vert$} Monograph Publication},'' Jan. 2024, [Online; accessed 2. Jan. 2024]. [Online]. Available: \url{https://www.ijsrp.org/monograph/detection-of-NPK-and-pH-components-soil-preface.php}
\BIBentrySTDinterwordspacing

\bibitem{s23021007}
\BIBentryALTinterwordspacing
L.~Silva, F.~Rodríguez-Sedano, P.~Baptista, and J.~P. Coelho, ``The digital twin paradigm applied to soil quality assessment: A systematic literature review,'' \emph{Sensors}, vol.~23, no.~2, 2023. [Online]. Available: \url{https://www.mdpi.com/1424-8220/23/2/1007}
\BIBentrySTDinterwordspacing

\bibitem{JONES202036}
\BIBentryALTinterwordspacing
D.~Jones, C.~Snider, A.~Nassehi, J.~Yon, and B.~Hicks, ``Characterising the digital twin: A systematic literature review,'' \emph{CIRP Journal of Manufacturing Science and Technology}, vol.~29, pp. 36--52, 2020. [Online]. Available: \url{https://www.sciencedirect.com/science/article/pii/S1755581720300110}
\BIBentrySTDinterwordspacing

\bibitem{VERDOUW2021103046}
\BIBentryALTinterwordspacing
C.~Verdouw, B.~Tekinerdogan, A.~Beulens, and S.~Wolfert, ``Digital twins in smart farming,'' \emph{Agricultural Systems}, vol. 189, p. 103046, 2021. [Online]. Available: \url{https://www.sciencedirect.com/science/article/pii/S0308521X20309070}
\BIBentrySTDinterwordspacing

\bibitem{Slob2022Sep}
N.~Slob and W.~Hurst, ``{Digital Twins and Industry 4.0 Technologies for Agricultural Greenhouses},'' \emph{Smart Cities}, vol.~5, no.~3, pp. 1179--1192, Sep. 2022.

\bibitem{article}
A.~Nasirahmadi and O.~Hensel, ``Toward the next generation of digitalization in agriculture based on digital twin paradigm,'' \emph{Sensors}, vol.~22, p. 498, 01 2022.

\bibitem{10.1016/j.procs.2022.10.118}
\BIBentryALTinterwordspacing
R.~N. Madeira, P.~A. Santos, O.~Java, T.~Priebe, E.~Gra\c{c}a, E.~S\'{a}rk\"{o}zi, B.~Asprion, and R.~P.-B. G\'{o}mez, ``Towards digital twins for multi-sensor land and plant monitoring,'' \emph{Procedia Comput. Sci.}, vol. 210, no.~C, p. 45–52, jan 2022. [Online]. Available: \url{https://doi.org/10.1016/j.procs.2022.10.118}
\BIBentrySTDinterwordspacing

\bibitem{article3table}
N.~Peladarinos, D.~Piromalis, V.~Cheimaras, E.~Tserepas, R.~Munteanu, and P.~Papageorgas, ``Enhancing smart agriculture by implementing digital twins: A comprehensive review,'' \emph{Sensors}, vol.~23, p. 7128, 08 2023.

\end{thebibliography}
\nocite{*}

\end{document}